**RESEARCH ARTICLE**

# Adaptive Optimizable Gaussian Process Regression Linear Least Squares Regression Filtering Method for SEM Images

DOMINIC CHEE YONG ONG[1], (Student Member, IEEE), IKSAN BUKHORI[2],
KOK SWEE SIM[1], (Senior Member, IEEE), AND KOK BENG GAN[3]
[1]Faculty of Engineering and Technology, Multimedia University, Melaka 75450, Malaysia
[2]Department of Electrical Engineering, Faculty of Engineering, President University, Bekasi 17550, Indonesia
[3]Department of Electrical, Electronic and Systems Engineering, Faculty of Engineering and Built Environment, Universiti Kebangsaan Malaysia, Bangi 43600, Malaysia

Corresponding author: Kok Swee Sim (sksbg2022@gmail.com)

This work was supported by Multimedia University.

**ABSTRACT** Scanning Electron Microscopy (SEM) images often suffer from noise contamination, which degrades image quality and affects further analysis. This research presents a complete approach to estimate their Signal-to-Noise Ratio (SNR) and noise variance (NV), and enhance image quality using NV-guided Wiener filter. The main idea of this study is to use a good SNR estimation technique and infuse a machine learning model to estimate NV of the SEM image, which then guides the wiener filter to remove the noise, providing a more robust and accurate SEM image filtering pipeline. First, we investigate five different SNR estimation techniques, namely Nearest Neighbourhood (NN) method, First-Order Linear Interpolation (FOL) method, Nearest Neighbourhood with First-Order Linear Interpolation (NN+FOL) method, Non-Linear Least Squares Regression (NLLSR) method, and Linear Least Squares Regression (LSR) method. It is shown that LSR method to perform better than the rest. Then, Support Vector Machines (SVM) and Gaussian Process Regression (GPR) are tested by pairing it with LSR. In this test, the Optimizable GPR model shows the highest accuracy and it stands as the most effective solution for NV estimation. Combining these results lead to the proposed Adaptive Optimizable Gaussian Process Regression Linear Least Squares Regression (AO-GPRLLSR) Filtering pipeline. The AO-GPRLLSR method generated an estimated noise variance which served as input to NV-guided Wiener filter for improving the quality of SEM images. The proposed method is shown to achieve notable success in estimating SNR and NV of SEM images and leads to lower Mean Squared Error (MSE) after the filtering process.

## I. INTRODUCTION
### A. BACKGROUND
Scanning Electron Microscope (SEM) imaging serves vital functions in all three fields of materials science, nanotechnology and biological research. A material's structure together with its surface can be displayed through highly detailed images delivered by this technology [1].

The associate editor coordinating the review of this manuscript and approving it for publication was Sudhakar Radhakrishnan.

SEM functions differently from optical microscopy by directing high-energy electron beams which repel signals from the sample surface involving primary electrons, second electrons and backscattered electrons. High-resolution images emerge from processed signals obtained by the detection system in the form of Secondary Electron (SE detector). The analysis of nanometer-scale features including grain boundaries together with defects and topographical attributes highly depends on SEM functionality [2].

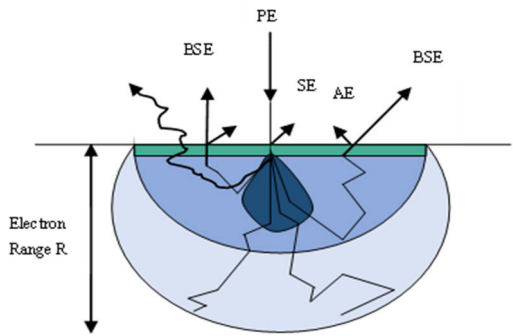

**FIGURE 1.** Electron-specimen interaction types and information depths in SEM imaging [3].

SEM represents an advanced imaging device that includes an electron gun alongside devices for lensing and scanning coils with associated signal detectors. The device releases 1–50 keV electron beams with high energy levels for specimen interaction resulting in image formation. The electrons from the electron gun are typically called Primary Electrons (PE). Once these electrons hit the specimen, the electrons are repelled as Backscattered Electrons (BSE), Secondary Electrons (SE), and Auger Electrons (AE) as shown in Fig. 1

In the industry, the SEM images are obtained from the SE via SE detector as shown in Fig. 2. Since the SE detector is the one which detects signals and forms the images, in this study we focus on the noises on these SE images. In this study, we focus on the Gaussian noise for two reasons. First, since these images came from SE detector, Poisson noise is not present here. Second, many literatures suggested that Gaussian noise analysis is often a good approximation in SEM images. For example, Goldstein in [3] mentioned that the statistical noise inherent in a detector system is commonly modeled using Gaussian distribution. Barret and Mayers also mentioned that noise generated from a lot of independent processes tend to be Gaussian by central limit theorem [4]. Similar argument was also posed by Rheimer in [5] in which he said that due to the large number of random effects, noise distribution in SEM imaging systems is commonly assumed to be Gaussian. Similarly, Dong et al. in [6] also assumed Gaussian noise as the standard baseline for image restoration models. Due to these arguments, in this study Gaussian noise in the SEM images are assumed.

The acquisition process of SEM images might create considerable background noise which reduces image quality. Three major noise sources including environmental disturbances and slow scan rates together with electronic interferences degrade image quality while preventing accurate analysis according to [7]. Noise reduction techniques need to be implemented since they determine SEM imaging reliability for both scientific research and industrial usage.

SEM evaluation of image quality depends strongly on Signal-to-Noise Ratio (SNR) measurements. The measurement technique evaluates how much meaningful signal stands compared to background noise. Better analysis requires higher SNR values because increased SNR produces images that have less noise and show better clarity [8]. SEM image SNR evaluation remains difficult because noise properties adapt to multiple factors including image resolution, contrast, and textural qualities of the image [9]. Machine learning represents one of the advanced methods that helps in achieving precise noise measurement. Support Vector Machines (SVM) and Gaussian Process Regression (GPR) are examples of such methods.

Various filtering approaches exist for boosting SEM images through noise reduction which maintains key image features. Early image filtering methods such as average, median, and Gaussian filters demonstrate poor results for maintaining image elements during their process of noise reduction [10]. The Wiener filter together with other adaptive filtering methods utilizes noise variance (NV) estimates to optimize its operational efficiency. Due to its image-specific noise modelling capabilities the Wiener filter executes filtering operations successfully without affecting fine image details.

In this research, a consolidated SEM image filtering system is developed through the combination of SNR estimation methods and adaptive filtering systems. We tested machine learning techniques, namely Support Vector Machines (SVM) and Gaussian Process Regression (GPR) to estimate the noise variance based on the estimated SNR values in this research. These estimates lead to an enhanced Wiener filter which increases SEM image clarity which allows for more accurate and reliable image analysis for scientific and industrial purposes.

### B. PROBLEM STATEMENT
SEM imaging process produces high-resolution imaging yet it suffers from noise issues that arise from detector imperfections as well as electron beam instability and environmental conditions and scan speed variations [11]. The noise causes deterioration of the image quality which hinders precise analysis in scientific and industrial fields. Numerous techniques have been developed to estimate the SNR of SEM images, ranging from traditional cross-correlation approaches, which require precise alignment of two images, to advanced models that use only a single image. Besides, the existing approaches for SNR estimation and noise variance estimation often struggle to offer precise adaptable solutions which results in below-average filtering results. Image analysis with human operators proves to be slow when dealing with noisy SEM images and traditional filters often find it hard to achieve satisfactory noise minimization while preserving image details simultaneously. In order to address these limitations, this research develops an integrated framework combining SNR estimation, noise variance estimation, and adaptive Wiener filtering to enhance SEM image quality to get a clearer SEM image after filtered.

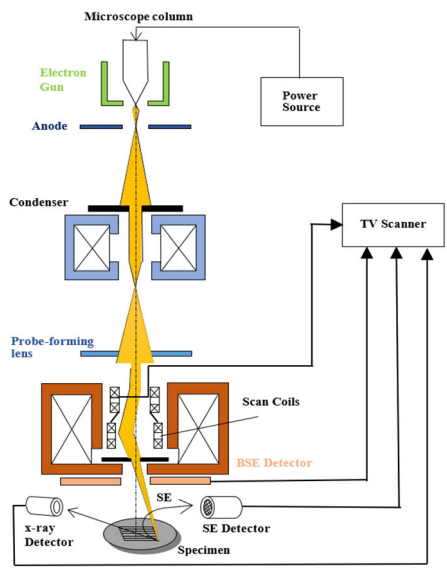

**FIGURE 2.** Working principle of SEM [17].

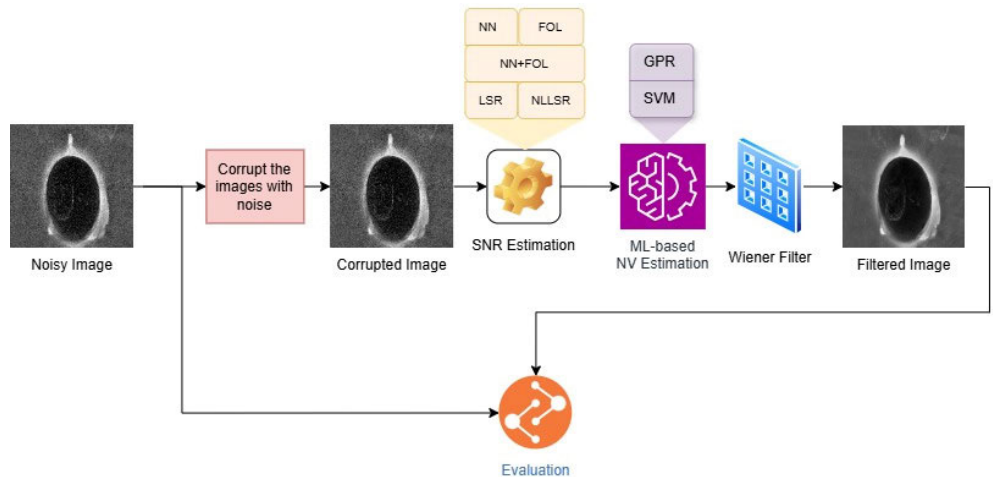

**FIGURE 3.** The design study to find the best combination of SNR estimation and machine learning-based NV estimation algorithm.

### C. PROPOSED APPROACH

This research uses adaptive filtering to minimize noise from SEM images. Gaussian and median filtering methods from traditional approaches prove ineffective for high-precision SEM by providing substandard results at detailed elimination. Wiener Filter provides an alternative solution for image denoising. In practice, the Wiener filter estimates the local mean and variance of the noisy image and then uses the provided noise variance to determine the extent of filtering. When the local variance is close to the noise variance, more smoothing is applied; when it's significantly higher, the filter preserves more of the original signal detail. This balance is what allows the Wiener filter to perform edge-preserving denoising.

The idea of this study is to introduce a new technique to feed this estimated NV to the Wiener Filter, which serves as a crucial input for adaptively controlling the filter response in relation to local image statistics. The advancement of machine learning techniques nowadays which enables high accuracy noise variance estimation as well as adaptive filtering may lead to superior outcomes than traditional approaches do in terms of SEM images filtering.

The research presents a systematic SEM image noise management pipeline which starts with SNR estimation to determine the most effective method for SEM images' SNR estimation. Five existing SNR estimation methods, which are NN method, FOL method, NN+FOL method, NLLSR method, and LSR method [12] are compared to evaluate their accuracy and reliability in estimating SNR values.

Next step is to estimate the noise variance of the SEM images. The noise variance estimation is performed using two algorithms which are called Support Vector Machines (SVMs) and Gaussian Process Regression (GPR). Different versions of SVM and GPR models are evaluated for this comparison.

Fig. 3 shows the pipeline of this study. The proposed approach is tested using SEM images obtained from experiments under uniform conditions utilizing one specific SEM machine. Original SEM images are corrupted with white Gaussian noise at varying noise levels to simulate real-world conditions, in order to test the performance of the proposed approach. The proposed methodology measures both classification and regression performance by using RMSE together with $R^2$ values. The proposed method shows its capability to enhance SEM images through decreased Mean Squared Error (MSE) values in its outcome.

The evaluations are based on RMSE and MSE and $R^2$ together with MAE measurements. As will be presented later, the evaluation shows that LSR is the best SNR estimator, while GPR outperforms the SVM. The proposed model named Optimizable GPR model utilized this result. The noise variance obtained from this model is then used to guide a a wiener filter, forming a variant of Wiener filter we call NV-guided Wiener Filter. This novel variant of Wiener Filter is used in conjunction with proposed system named Adaptive Optimizable Gaussian Process Regression Linear Least Squares Regression Filtering (AO-GPRLLSR), as shown in Fig. 4. The adaptive filtering method adjusts its operation based on noise prediction levels to properly suppress noise disturbances without degrading SEM image features.

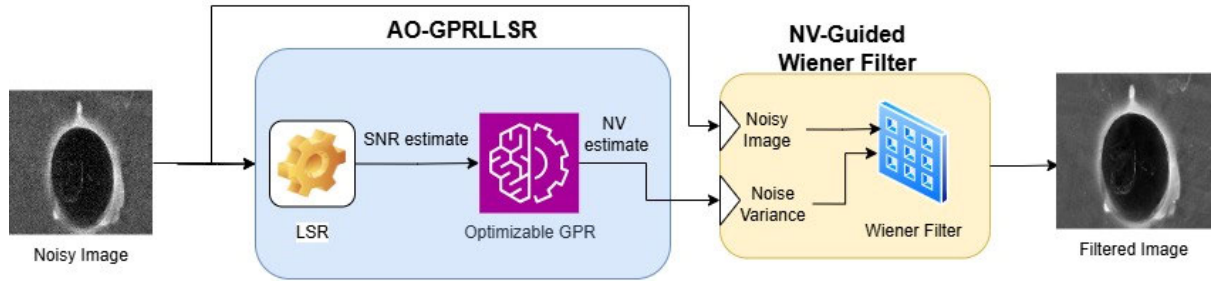

**FIGURE 4.** Proposed Approach: Adaptive Optimizable Gaussian Process Regression Linear Least Squares Regression Filtering (AO-GPRLLSR).

### D. CONTRIBUTIONS

The research introduces a novel integrated SEM image noise filtering approach that uses advanced SNR estimation and machine learning noise variance estimation and adaptive filtering techniques. This research implementation includes a complete analysis of five SNR evaluation methods including NN method, FOL method, NN+FOL method, NLLSR method, and LSR method. The experimental results have confirmed the LSR method as the leading option to accurately measure SNR in SEM images. The discovery provides necessary guidelines for choosing suitable SNR estimation methods in high-resolution imaging processes.

Furthermore, the research presents a noise variance estimation method using different regression models (SVM and GPR models). Experiment outcomes show that Optimizable GPR delivers optimal noise variance estimation when evaluated using RMSE and MSE, $R^2$ and MAE performance metrics. The establishment of Optimizable GPR to provide the best results for noise variance estimation would enhance precision in SEM image processing.

The final contribution is the development of a novel Adaptive Optimizable Gaussian Process Regression Linear Least Squares Regression (AO-GPRLLSR) Filtering method that dynamically adjusts to noise estimates to enhance SEM image quality. The proposed adaptive Wiener filter surpasses conventional filtering techniques since it manages to reduce SEM image noise effectively while maintaining the crucial structural elements present in SEM images.

This research integrates novel development into an integrated process which both elevates SEM image analysis reliability and creates a foundational platform for future automated noise classification, estimation, and filtering research. These proposed methods introduce new capabilities to materials science and nanotechnology research as well as biomedical research through their ability to improve image quality in SEM images.

In the following sections, we present a detailed review of related work, describe our methodology, discuss experimental results, provide the proposed method's challenges and possible improvements and make a conclusion on this research.

## II. RELATED WORKS
### A. SNR ESTIMATION

Signal-to-Noise Ratio (SNR) evaluation helps to determine image quality while assessing both SEM image noise and SEM image clarity [13], [14]. The traditional SNR assessment method depends on spectral analysis in addition to statistical measures of pixel intensities. For example, the assessment of signal and noise frequency domain components utilizes Power Spectral Density (PSD) as a frequently employed technique for SNR estimation [15]. However, these methods remain insufficient to process diverse noise patterns which frequently appear in SEM images.

Therefore, advanced, comprehensive, and reliable models have become a solution to tackle the difficulties encountered in SNR estimation [16]. Hence, this research analyses the effectiveness of five existing SNR estimation techniques including NN method, FOL method, NN+FOL method, NLLSR method and LSR method to identify the best method. These five existing methods determine the SNR estimation value based on autocorrelation function (ACF) curve of the noisy SEM image. Fig. 3 shows the Autocorrelation function (ACF) curve of a single SEM image corrupted with White Gaussian noise [17].

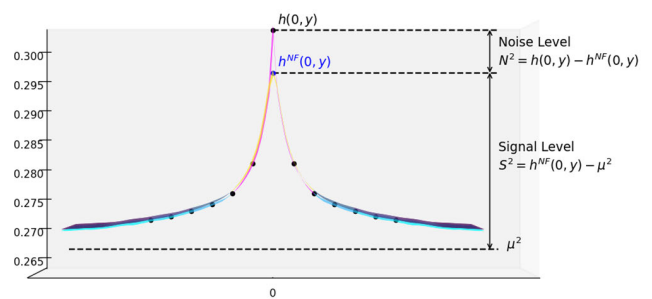

**FIGURE 5.** Autocorrelation function (ACF) curve [17].

Equation (1) shows the general SNR estimation equation. The SNR numerator value depends on the corrupted SEM image's mean value, $\mu^2$ subtracted from original image's ACF peak, $h^{NF}(0,y)$, and difference between the noisy ACK peak value $h(0,y)$, and $h^{NF}(0,y)$. The value of $h(0,y)$ and $\mu^2$ can be directly obtain from the image's ACF curve while the value of $h^{NF}(0,y)$ has to be estimated in order to obtain the SNR value. Hence, different methods have been established to estimate $h^{NF}(0,y)$. These methods include NN method and FOL method, NN+FOL method, NLLSR method and LSR method as shown in (2) provides the formula to calculate dB values of SNR.

$$SNR = \frac{h^{NF}(0,y) - \mu^2}{h(0,y) - h^{NF}(0,y)} \quad (1)$$

$$SNR(dB) = 20\log[10(SNR)] \quad (2)$$

### 1) NEAREST NEIGHBOURHOOD (NN) METHOD
The Nearest Neighbourhood (NN) method, introduced by Kamel and Sim in [18], is one of the simplest approaches for SNR estimation. It estimates SNR based on the closest points to the x-axis of the noisy autocorrelation function (ACF) peak, $h(0,y)$. The SNR estimation is computed using the (3) and (4), where $h(0,y)$ represents the ACF peak value, $h(1,y)$ and $h(-1,y)$ are the nearest ACF points and $u^2$ is the estimated noise variance.

$$SNR = \frac{h(1,y) - u^2}{h(0,y) - h(1,y)} \quad (3)$$

$$SNR = \frac{h(-1,y) - u^2}{h(0,y) - h(-1,y)} \quad (4)$$

### 2) FIRST-ORDER LINEAR INTERPOLATION (FOL) METHOD
The First-Order Linear Interpolation (FOL) method [18], also proposed by Sim, refines SNR estimation by considering four points, which are two points on each side of $h(0,y)$. By interpolating between these points, a more accurate estimate of the peak is obtained. The SNR estimation is calculated as (5), where $h^{int}(0,y)$ represents the noisy ACF peak at $h(0,y)$.

$$SNR = \frac{h^{int}(0,y) - u^2}{h(0,y) - h^{int}(0,y)} \quad (5)$$

### 3) NEAREST NEIGHBOURHOOD WITH FIRST ORDER LINEAR INTERPOLATION (NN+FOL) METHOD
NN+FOL method combines the strengths of NN and FOL by averaging their results to provide a more robust SNR estimate, also proposed by Sim [18]. The estimated ACF peak is computed as (6), (7) and (8).

$$h^{nn+int}(0,y) = \frac{h^{nn}(0,y) + h^{int}(0,y)}{2} \quad (6)$$

$$h^{nn+int}(0,y) = \frac{3[h(1,y)] - h(2,y)}{2} \quad (7)$$

$$h^{nn+int}(0,y) = \frac{3[h(-1,y)] - h(-2,y)}{2} \quad (8)$$

The SNR estimation is shown in (9), where $h^{NF}(0,y)$ represents the noisy ACF peak estimation using NN and FOL. This hybrid method aims to enhance precision by leveraging both approaches.

$$SNR = \frac{h^{NF}(0,y) - \mu^2}{h(0,y) - h^{NF}(0,y)} \quad (9)$$

### 4) NON-LINEAR LEAST SQUARES REGRESSION (NLLSR) METHOD
NLLSR method, proposed by Sim & Norhisham in 2016, models the ACF curve using a non-linear exponential relationship [19]. Equation (10) shows the estimated original image's peak, $\hat{Y}^{NLLSR}$.

$$\hat{Y}^{NLLSR} = \hat{Y} = \alpha\varepsilon\prod_{k=1}^{M}(X^k)^{\beta_k} \quad (10)$$

The SNR estimation is shown in (11), where the $\hat{Y}^{NLLSR}$ is the estimated original image's peak.

$$SNR = \frac{\hat{Y}^{NLLSR} - u^2}{h(0,y) - \hat{Y}^{NLLSR}} = \frac{[\alpha\varepsilon\prod_{k=1}^{M}(X^k)^{\beta_k}] - u^2}{h(0,y) - [\alpha\varepsilon\prod_{k=1}^{M}(X^k)^{\beta_k}]} \quad (11)$$

### 5) LINEAR LEAST SQUARES REGRESSION (LSR) METHOD
LSR method, also proposed by Sim & Norhisham applies a linear regression model to estimate the ACF peak [12]. This method assumes that the ACF follows a linear trend, allowing for a simplified and effective noise estimation. Eq. (12) shows the estimated original image's peak by LSR method, $h^{LSR}$, while (13) shows the error formula of the $h^{LSR}$.

$$h^{LSR}(0,y) = \hat{Y} = \alpha + BX + \varepsilon \quad (12)$$

$$\varepsilon = \frac{h(0,y) - h(1,y)}{2} \quad (13)$$

The SNR estimation is shown in (14), where the $h^{LSR}(0,y)$ is the estimated of the original image's peak.

$$SNR = \frac{h^{LSR}(0,y) - u^2}{h(0,y) - h^{LSR}(0,y)} = \frac{(\alpha + BX + \varepsilon) - u^2}{h(0,y) - (\alpha + BX + \varepsilon)} \quad (14)$$

## B. NOISE VARIANCE ESTIMATION
The estimation of noise variance plays an essential role in images noise reduction since it provides the measurement for image noise levels [20], [21]. The estimation of noise variance through traditional models requires statistical analysis of homogeneous image sections for determination. Equation (15) shows the traditional noise variance formula which uses the difference between original ACF peak, $h^{NF}(0,y)$ and noisy ACF peak, $h(0,y)$ divided by the image size.

$$Noise\ Variance = \frac{h(0,y) - h^{NF}(0,y)}{Image\ Size} \quad (15)$$

However, the existing methods do not work effectively to adjust their performance when faced with complex or spatially evolving noise patterns which frequently appear in SEM images. Hence, modifications are needed to address this problem.

The noise distribution modeling capabilities of Support Vector Machines (SVMs) and Gaussian Process Regression (GPR) techniques have made these techniques popular in machine learning applications. SVMs find broad application in regression tasks because they effectively analyze high-dimensional datasets. Research that employed SVMs for microscopy image noise variance estimation yielded promising outcomes according to previous work [22]. An optimal performance requires proper selection of kernel functions along with hyperparameters according to [23]. In this research, several SVM models and GPR models were compared. These include Linear SVM, Quadratic SVM, Cubic SVM, Fine Gaussian SVM, Medium Gaussian SVM, Coarse Gaussian SVM, Rational Quadratic GPR, Squared Exponential GPR, Matern 5/2 GPR, Exponential GPR, Optimizable SVM and Optimizable GPR.

### 1) LINEAR SVM
Linear SVM uses a linear decision boundary. The linear decision boundary of Linear SVM provides effective results on linearly separable patterns despite its inability to detect intricate noise patterns.

$$f(x) = \omega' x + b \tag{16}$$

Equation (16) shows the decision boundary formula where $\omega$ is the weight vector, $x$ is he input feature vector and $b$ is the bias term.

### 2) QUADRATIC SVM
The second-degree polynomial kernel in Quadratic SVM provides limited non-linear. Equation (17) shows the model formula which allows it to capture more complex relationship between features compared to the Linear SVM.

$$f(x) = \omega' x + b \tag{17}$$

### 3) CUBIC SVM
Cubic SVM uses a third-degree polynomial kernel as shown in (18).

$$K(x, y) = (x'y + 1)^3 \tag{18}$$

### 4) FINE GAUSSIAN SVM
Fine Gaussian SVM implements the Gaussian Radial Basis Function (RBF) kernel with small sigma value to detect precise data features. The small bandwidth Gaussian kernel in Fine Gaussian SVM produces high sensitivity to local noise but makes it susceptible to overfitting. Equation (19) shows the equation of Fine Gaussian SVM, where $\sigma$ is a small-scale parameter.

$$K(x, y) = exp\left(-\frac{\|x - y\|^2}{2\sigma^2}\right) \tag{19}$$

### 5) MEDIUM GAUSSIAM SVM
Medium Gaussian SVM model also uses the Gaussian RBF kernel but with a moderate $\sigma$, providing a balance between complexity and generalization.

### 6) COARSE GAUSSIAN SVM
With a large $\sigma$, Coarse Gaussian SVM smooths out decision boundaries, making it less sensitive to noise but potentially missing fine details.

The probabilistic nature of GPR has become a robust substitution because it produces point estimates alongside confidence intervals [24]. GPR proves valuable for uncertain noise conditions that can be found in SEM images because of its probabilistic nature.

### 7) RATIONAL GAUSSIAN SVM
The Rational Quadratic kernel is a scale mixture of RBF kernels as shown in (20), where $\alpha$ controls the weighting of scales and $l$ is the length scale.

$$K(x, y) = exp\left(1 + \frac{\|x - y\|^2}{2\alpha l^2}\right)^{-\alpha} \tag{20}$$

### 8) SQUARED EXPONENTIAL GPR
The Squared Exponential kernel is the most widely used GPR kernel as shown in (21), where $\sigma^2$ is the variance parameter and $l$ is the length scale controlling smoothness of the model.

$$K(x, y) = \sigma^2 exp\left(-\frac{\|x - y\|^2}{2l^2}\right) \tag{21}$$

### 9) MATERN 5/2 SVM
Matern 5/2 kernel is more flexibility than the Squared Exponential kernel. Equation (22) shows the kernel equation of Matern 5/2 GPR model.

$$K(x, y) = \sigma^2 \left(1 + \frac{\sqrt{5}\|x - y\|^2}{l} + \frac{5\|x - y\|^2}{3l^2}\right) exp\left(-\frac{\sqrt{5}\|x - y\|}{l}\right) \tag{22}$$

### 10) EXPONENTIAL GPR
Exponential kernel is a special case of Matern kernel. Equation (23) shows the kernel equation. This kernel models fewer smooth functions.

$$K(x, y) = \sigma^2 exp\left(-\frac{\|x - y\|}{l}\right) \tag{23}$$

### 11) OPTIMIZABLE SVM

An Optimizable SVM model is used, leveraging Bayesian Optimization to determine the most suitable SVM kernel and hyperparameters [25]. Bayesian Optimization evaluates performance based on expected improvement, probability of improvement, lower confidence bound, and standard deviation of the objective function [26].

For Bayesian Optimization [27], to determine the best SVM kernel and SVM kernel parameters, it assesses a point x's performance using the posterior distribution function, Q. The performance or parameters are based on the expected improvement, probability of improvement, lower confidence bound, time taken (per second) and standard deviation of the posterior objective function (plus).

Equation (24) shows the expected improvement, $EI(x, Q)$ equation, which $x_{best}$ is the location of lowest posterior mean and $\mu_Q (x_{best})$ is the lowest value of posterior mean.

$$EI(x, Q) = E_Q[\max(0, \mu_Q (x_{best}) - f(x))] \quad (24)$$

While (27) displays the $v_Q(x)$ formula in Eq. (26), a simplified form of (25) is given by (26). The probability of improvement is expressed as $PI(x, Q)$; the estimated standard deviation of noise is represented by the margin parameter, m. Subsequently, the posterior standard deviation of the process at x is represented by $\sigma_Q(x)$, where $T$ represents the unit normal cumulative distribution function (CDF).

$$PI(x, Q) = P_Q(f(x) < \mu_Q (x_{best}) - m) \quad (25)$$
$$PI = \Phi(v_Q(x)) \quad (26)$$
$$v_Q(x) = \frac{\mu_Q (x_{best}) - m - \mu_Q(x)}{\sigma_Q(x)} \quad (27)$$

Equation (28) and (29) show the equation formation of lower confidence bound (LCB).

$$G(x) = \mu_Q (x) - 2\sigma_Q(x) \quad (28)$$
$$LCB = 2\sigma_Q (x) - \mu_Q(x) \quad (29)$$

Time taken (per second) is used to calculate the time for SVM for dealing with the range of points, which can improve the time taken after obtaining per second. The formula of per second is shown in (30) which $EIpS(x)$ represents expected improvement per second and $\mu_S(x)$ represents the posterior mean of the time taken for the process.

$$EIpS(x) = \frac{EI_Q(x)}{\mu_S(x)} \quad (30)$$

Standard deviation of the posterior objective function (plus) equation as shown in (31) is used to avoid the over-exploitation of the model, where $\sigma$ is the posterior standard deviation of additive noise and $\sigma_F(x)$ is the standard deviation of the posterior objective function at $x$. Equation (32) is the condition of plus equation, where $t_\sigma$ represents the exploration ratio.

$$\sigma_Q^2 (x) = \sigma_F^2 (x) + \sigma^2 \quad (31)$$
$$\sigma_F(x) < t_\sigma \sigma \quad (32)$$

### 12) OPTIMIZABLE GPR

Optimizable GPR follows a similar Bayesian Optimization process to select the best kernel and hyperparameters [28]. The goal is to minimize prediction error using metrics like RMSE and log-likelihood. Equation (33) shows the posterior mean function of this model and (34) shows the variance of this model.

$$\mu(x) = K(x, X)' K(X, X)^{-1} Y \quad (33)$$
$$\sigma^2(x) = k(x, x) - K(x, X)' K(X, X)^{-1} K(x, X) \quad (34)$$

In the experimental section, the Optimizable SVM and Optimizable GPR models will be directly compared to assess their effectiveness in noise variance estimation. Key evaluation metrics such as RMSE, MSE, $R^2$, and MAE will be applied to determine the superior model for noise characterization in SEM images.

### C. IMAGE FILTERING

Image filtering methods are used to improve SNR of the grayscale image and, consequently, enhance its clarity [29] [30]. Mean Square Error (MSE) is the formula to calculate the efficiency of the filtering methods, which are shown in (35). $I(i, j)$ is the pixel value of the original clean image at position (i, j), $\hat{I}(i, j)$ is pixel value of the filtered image at position (i, j), and M×N is the dimensions of the image (number of rows × number of columns). A low MSE reflects a good filtering method [31].

$$MSE = \frac{1}{M \times N} \sum_{i=1}^{M} \sum_{j=1}^{N} \left(I(i, j) - \hat{I}(i, j)\right)^2 \quad (35)$$

In the proceeding section. we will discuss about the Average filter, Median filter, Gaussian filter and Wiener filter.

### 1) AVERAGE FILTER

Average filter works well for eliminating noise in images that contains salt and pepper, but it is not suitable for removing Gaussian white noise, since the degree of intensity of this type of noise is evenly distributed according to the Gaussian bell shape [32]. To get the mean pixel intensity, the Average filter essentially takes the average of the center pixel and the pixels surrounding it. The intensity mean of this pixel is then re-distributed throughout the filter window or mask. Thus, the average filter is a member of the linear low-pass class. The two dimensions of the filter window with a (3 × 3) resolution are shown in Fig. 4.

Based on Fig. 4, the mean intensity of an image is obtained by adding all of the pixel intensities and dividing the result by the window's size. Equation (36) shows the general equation to obtain the mean of the average filter, where X is the intensity of the pixel located in the x-coordinate and Y is the intensity of the pixel located in the y-coordinate. Fig. 5 shows the average out pixel

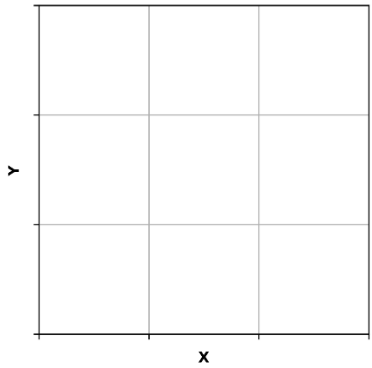

FIGURE 6. Filter window of size (3 × 3).

intensity within the window size of (3 × 3).

$$Pixel\ Mean\ (\mu) = \frac{1}{(XY)} \sum_{N=1}^{XY} A_N \quad (36)$$

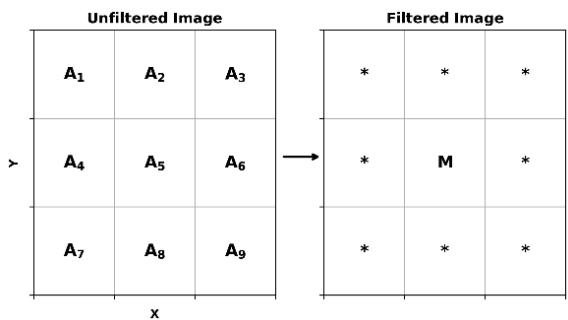

FIGURE 7. Unfiltered image and average filtered image.

This filter's disadvantage arises when there are extremely high or extremely low pixel intensities within the filter window, which might cause the filtered image to become blurry [33]. To clarify it, the extreme pixel intensity value may have an impact on the average filter performance [34]. Therefore, a grayscale image with white Gaussian noise that randomly modifies the pixel intensity value between 0 and 255 different black and white intensities is unsuitable for the Average filter.

Previously, in Gonzales & Woods highlight, many kinds of noise filters have been designed to deal with different kinds of noise [35]. The simplest and most widely used filter to remove noise from an image is the average filter. As implied by the name, this filter averages the intensity of each pixel inside a given window mask by taking the mean of the center and surrounding pixels. Because the intensity of the pixels differs significantly from their initial intensity, this eventually led to the filtered image being blurry.

### 2) MEDIAN FILTER

The non-linear digital filter known as the median filter is effective in eliminating salt and pepper noise [36]. Median filters, in contrast to Average filters, only take into account the median size of the mask, kernel, or filter window [37]. The first step is numerically rearranging the filter's pixel intensity from lowest to greatest. Subsequently, the Median filter will select the new pixel intensity value based on the median size of the filter mask [38]. The fourth highest pixel intensity will be selected by the median filter, for instance, if the median size is four [39]. Next, the filtered mask's new pixel intensity, represented by M, is relocated as the center of the mask. Fig. 6 shows the filter mask of unfiltered image and Median filtered image. Equation (37) is used to obtain the median size of the filter mask, where X is the intensity of the pixel located in the x-coordinate and Y is the intensity of the pixel located in the y-coordinate.

$$Median = \frac{XY}{2} \quad (37)$$

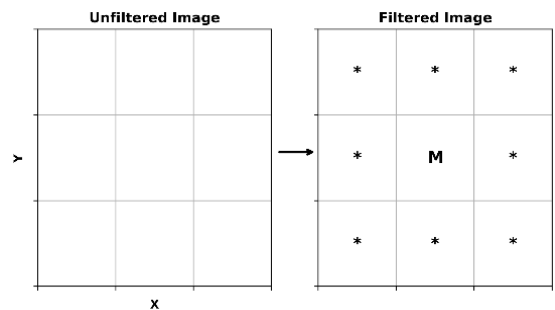

FIGURE 8. Unfiltered image and median filtered image.

Extremely high or low pixel intensity in the unfiltered image has no effect on the median filter. Because of this, it can maintain the image's sharp edge [40]. This suggested that, in contrast to the Average filter, the Median filter can avoid visual blurring [34]. The disadvantage is that the median filter might potentially eliminate the true pixel intensity of neighbouring pixels. It is therefore unsuitable for eliminating white Gaussian noise with evenly distributed intensity. To eliminate the white Gaussian noise, the median size must be chosen adaptively. Hence, there were many improved median filters developed based on the concept of median filter.

Over the past few decades, numerous improvements to the median filter have been proposed. The Adaptive Median filter was created in 2003 to improve microarray pictures [41]. In order to eliminate impulsive noise from color images, Adaptive varying window size Recursive Weighted Median Filter (ARWMF) was created in 2007 [42]. In order to improve magnetic resonance (MR) images, Arastehfar et al. developed a new method known as the decision-based median (DBM) filter, which they presented as an improvement to the median filter [43]. This approach is predicated on the previous determination of whether a single pixel can be used to compute the median. Moreover, Juneja et al. suggested the Improved Adaptive Median filter for noise caused by salt and pepper in 2009 [44]. Rather than filtering every pixel in the image, it recognizes the corrupted pixels and only applies the filter to those. In addition, Juneja et al. presented a few hybrid filters for improved edge preservation. The way this hybrid filter operates is by substituting the median of the 4 neighborhoods only for the pixel value of a point z. However, it is limited to noise in the salt and pepper category and is not effective in images with high noise density or significant noise variance. Its ability to eliminate white Gaussian noise from grayscale and SEM images is also lacking.

### 3) GAUSSIAN FILTER

A low-pass filter called a Gaussian filter reduces the image's high frequency component [45]. This filter is a member of the linear smoothing filter class. As seen in Fig. 7, the Gaussian filter applies a variable kernel or mask size with a bell-like Gaussian density function shape.

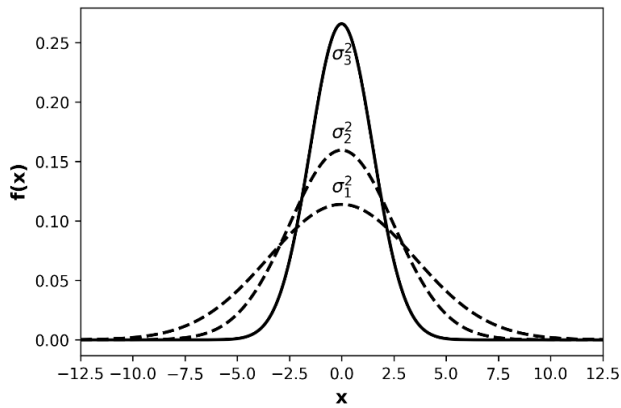

**FIGURE 9.** Gaussian density function with different noise variance, $\sigma^2$ [46].

In Fig. 7, the standard deviation, $\sigma$ in a SEM image corresponds to the noise variance, $\sigma^2$ of white Gaussian noise. Equation (38) formulates the Gaussian filtered image based on the characteristics of this distribution.

$$G(x, y) = \frac{1}{2\pi\sigma^2} e^{-\left[\frac{(x^2+y^2)}{2\sigma^2}\right]} \quad (38)$$

Equation (38) shows that $G(x, y)$ is the image's value that has been Gaussian filtered. The white noise variance, $\sigma^2$ value determines how accurate and smooth the filtering is. When it comes to eliminating white Gaussian noise from SEM pictures, the Gaussian filter shows excellent performance. On the other hand, the Gaussian filter performs terribly at eliminating salt and pepper noise. In practical case, this filter is physically unrealizable [47].

In the fields of computer vision and image processing, Gaussian filtering has also been thoroughly researched [48]. The adaptive gaussian filter was proposed by Deng et al. in 1993. According to the findings, an adaptive Gaussian filter always results in a reduced mean square error (MSE) than an image processed by a non-adaptive Gaussian filter, or the original Gaussian filter [48]. It also produces less distortion to the edges.

### 4) WIENER FILTER

The Wiener filter is a member of the optimal linear filter class. The capacity of the Wiener filter to reduce the mean square error (MSE) between the original and corrupted image is its primary benefit [49]. After filtering out the white noise, the Wiener filter can simultaneously eliminate visual blur of the image [50]. As a result, stationary processes involving additive white noise frequently use this filter.

$$W(x, y) = \frac{P_s(x, y)}{P_s(x, y) + \sigma^2} \quad (39)$$

Equation (39) describes how the Wiener filter performs image filtering based on additive white noise. $W(x, y)$ is the value of the Wiener filtered image, where x and y are the kernel dimension. The power spectrum, or $P_s(x, y)$, is derived from the original signal's Fourier Transform, where $\sigma^2$ stands for white noise variance. The original image's autocorrelation function, not the corrupted image, is what the Fourier Transform of the original signal refers to.

The Wiener filter effectively eliminates white noise from the SEM image in terms of efficiency. Nevertheless, the Wiener filter does not focus on improvement of image quality because its primary focus is on decreasing the mean square error (MSE). This indicates that the filter's ability to increase the signal-to-noise ratio (SNR) is not very good. However, if the goal is limited to eliminating white noise from the SEM image and does not involve any further SEM image enhancement, the Wiener filter may be taken into consideration to filter the image.

Wiener filters are frequently used to reduce different kinds of noise present in images [35]. The Wiener filter is a type of optimum linear filter that reduces the corrupted images' mean square error (MSE). Arazm et al. developed the local adaptive weighted Wiener filter (LAWWF), a SEM image filtering technique in 2017 [51]. This method is then compared with original wiener filter and Wiener filter Adaptive Noise Wiener (ANW) filter which was developed by Sim et al. in 2016. The PSNR of LAWWF is higher compared to ANW filter and original wiener filter [51].

On the other hand, a white noise removal technique for grayscale and color images was also proposed by Harikrishnan et al. in 2013, and it was based on the fuzzy logic concept [52]. Higher intensity noise is primarily removed by this filter. However, processing was a little slow and required a complex computation. This method likewise yields a low MSE value and slightly high PSNR. Then, for filtering images corrupted with white Gaussian noise, Vijendran et al. developed Rao-Blackwellized Particle Filtering (RBPF), a combination of a particle filter and a Kalman filter [53]. Particle filters are used to determine discrete states, while Kalman filters are used to calculate the distribution of continuous states. This method is applied to eliminate white Gaussian noise with high intensity [53]. However, the quantity of particles that needed to be eliminated required complicated calculation.

**TABLE 1.** Summary of the filtering methods.

| Image filtering methods | Effectiveness | Advantages | Disadvantages |
|---|---|---|---|
| Average filter | Works well for eliminating noise in images that contain salt and pepper, but it is not suitable for removing Gaussian white noise. | • Simple and widely used for noise removal.<br>• Computes the mean intensity within a filter window. | • Causes image blurring.<br>• Ineffective for Gaussian noise due to the uniform distribution of noise intensity [32]. |
| Median filter | Effective in eliminating salt and pepper noise but unsuitable for eliminating white Gaussian noise. | • Preserves edges better than the Average filter [34].<br>• Less affected by extreme pixel intensities. | • May alter neighboring pixel intensities.<br>• Requires adaptive modifications to handle white Gaussian noise effectively. |
| Gaussian filter | It shows excellent performance but terribly at eliminating salt and pepper noise [47]. | • Uses a Gaussian-weighted kernel to smooth images.<br>• Preserves intensity variations better than the Average filter. | • Struggles with real-world adaptive filtering scenarios.<br>• Does not perform well for salt-and-pepper noise. |
| Wiener filter | Best for removing additive white Gaussian noise and improving SNR [50]. | • Minimizes Mean Square Error (MSE) between original and filtered images [50].<br>• Adapts to noise power spectrum, maintaining image clarity. | • Primarily focuses on noise reduction rather than overall image enhancement.<br>• Requires accurate noise variance estimation for optimal performance. |

Table 1 shows the summary of the filtering methods, which Wiener filter is most effective for removing additive white Gaussian noise while improving SNR, making it the best choice for this research.

As argued in [54], adaptive filtering technique is often preferable in removing noise, since the filter can adapt to the changing noise condition. Hence, in this study, noise variance estimation is to be integrated with Wiener filter, forming an NV-guided Wiener Filtering technique to make it more adaptive to the source image's noise level.

## III. METHODOLOGY

Our proposed workflow for SEM image noise management consists of three key stages: SNR estimation, noise variance estimation, and image enhancement through Wiener filtering. Each stage is carefully designed to ensure accuracy and efficiency, leveraging machine learning models and adaptive filtering techniques for optimal performance. Each of these stages are implemented in a Graphical User Interface (GUI) with the general workflow as shown in Fig. 10.

### A. SIGNAL TO NOISE RATIO ESTIMATION

The assessment of image quality requires Signal-to-Noise Ratio (SNR) estimation as its initial step which also serves as the base for noise variance evaluation. This research, five existing SNR estimation methods are compared, which are NN method, FOL method, NN+FOL method, NLLSR method, and LSR method to determine the most suitable approach for SEM images. Experimental results indicate that the LSR method outperforms the others in terms of accuracy and computational efficiency, making it the preferred choice for this study.

The LSR method estimates SNR by fitting a linear model to the intensity distributions of SEM images, effectively capturing the ratio of signal power to noise power. This approach provides a robust, data-driven estimation of SNR, which serves as the input for subsequent noise variance estimation.

### B. NOISE VARIANCE ESTIMATION

The precision of noise variance calculation remains vital to achieve effective noise reduction because it evaluates the quantity of image noise and enables adaptive filtering. At first, the SNR is estimated using the LSR method, providing a reliable noise characterization metric. After that, machine learning models, specifically Support Vector Machines (SVM) and Gaussian Process Regression (GPR), are utilized to estimate noise variance based on the obtained SNR values.

The selected SVM model suits regression tasks with numerous dimensions and helps to decrease forecast errors effectively. The model develops a connection between SNR values and noise variance by building effective relationships between image quality changes. To determine optimal kernel function and hyperparameters, Bayesian Optimization employs five different criteria such as expected improvement, probability of improvement, lower confidence bound, time taken (per second) and standard deviation of the posterior objective function (plus) for fine-tuning the Optimizable SVM model. The system selects the SVM Cubic kernel because it produces optimal performance results during noise variance estimation.

In addition to SVM, the research also uses GPR models. GPR stands as a reliable tool for noise variance management because it presents point estimates together with confidence intervals when dealing with SEM image uncertainties.

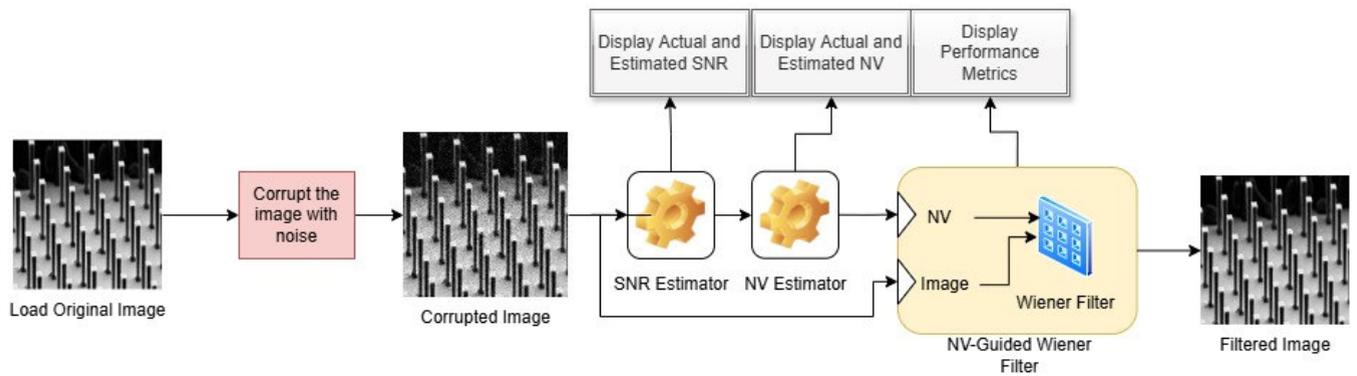

**FIGURE 10.** Methodology as implemented in GUI.

GPR uses its probabilistic method to find complex data relationships between SNR and noise variance which produces better predictive outcomes.

There are 12 models being compared in this research. These include Linear SVM, Quadratic SVM, Cubic SVM, Fine Gaussian SVM, Medium Gaussian SVM, Coarse Gaussian SVM, Rational Quadratic GPR, Squared Exponential GPR, Matern 5/2 GPR, Exponential GPR, Optimizable SVM and Optimizable GPR. The best-performing model is determined by evaluating RMSE, MSE, $R^2$, and MAE. Among the tested GPR models, the Optimizable Gaussian Process Regression (GPR) model with LSR tuning is identified as the most effective approach, providing accurate noise variance estimates while preserving confidence intervals for uncertainty quantification. Hence, the noise variance estimated by Optimizable GPR LSR was then used to be the reference of the filter to for an adaptive filter.

### C. IMAGE ENHANCEMENT

The proposed workflow ends with adaptive Wiener filtering that minimizes image noise by autotuning structural information in SEM images. The Wiener filter distinguishes itself from conventional filtering techniques by activating adjustments in its filtering processes according to estimated noise variation levels. The filter effectively reduces noise because of its dynamic adjustment process that allows important image details to remain intact.

The Wiener filter tailors its filtering process through the noise variance data obtained from the optimized SVM and GPR models. The filter makes better use of noise variance data to prevent both smoothing effects that degrade microscopic structures and damaging preservation of crucial features. SEM images require these characteristics because they experience variations in noise across different imaging scenarios along with variation in scanning parameters.

The filtering system performs successive operations which improve image clarity stepwise. Each processing step of the filter includes a deconvolution operation which both utilizes noise variance assessment and image power spectral density analysis to guide the operation.

The method encompasses step-wise noise reduction of high-frequency elements that successfully maintain structural characteristics including textures and edges. The Wiener filter adapts its parameters persistently to decrease Mean Squared Error between noise-reduced images and estimated original image versions which lead to optimal results.

To test the Wiener filter operational effectiveness, quantitative and qualitative evaluation methods were used. The MSE equation is shown in (35). The filtering process achieves success when the MSE between noisy and filtered images demonstrates a substantial decrease. After that, visual assessments are performed to verify that the improved images preserve their structural quality while displaying enhanced clarity. Tests validate Wiener filtering as a precise method for enhancing SEM image quality through optimized noise variance evaluation for microscopy applications.

In order to analyze and compare the SNR estimation techniques and image filtering methods, a GUI application is used. Fig. 11 shows the design of the GUI.

A GUI has been built for interactive processing of SEM images to offer users a convenient interface that combines functions of SNR evaluation and image noise addition followed by enhancement operations. Users can view the staged workflow through a structured format provided by the GUI interface.

The system request begins by allowing users to select any clean SEM image from the available dataset showing both image and its Autocorrelation Function (ACF) curve. Users can understand image structural elements by analyzing the Autocorrelation Function curve. The user defines the noise variance as the controlling parameter for determining the amount of added noise to the image. The user chooses the amount of noise variance on the interface and hits ''Convert'' to apply the noise which produces a noisy SEM image together with its new ACF curve.

The user can press ''Calculate'' after generating the noisy image because this action starts the SNR calculation process using LSR parameters. The application presents both the estimated SNR value and several other relevant outputs for further analysis. The user proceeds to choose filtering options during the current phase of operation. The selected filtering methodology processes the corrupted image through the interface which results in the presentation of the output data. The last image output shows enhanced clarity because it contains reduced noise while maintaining all structural elements intact.

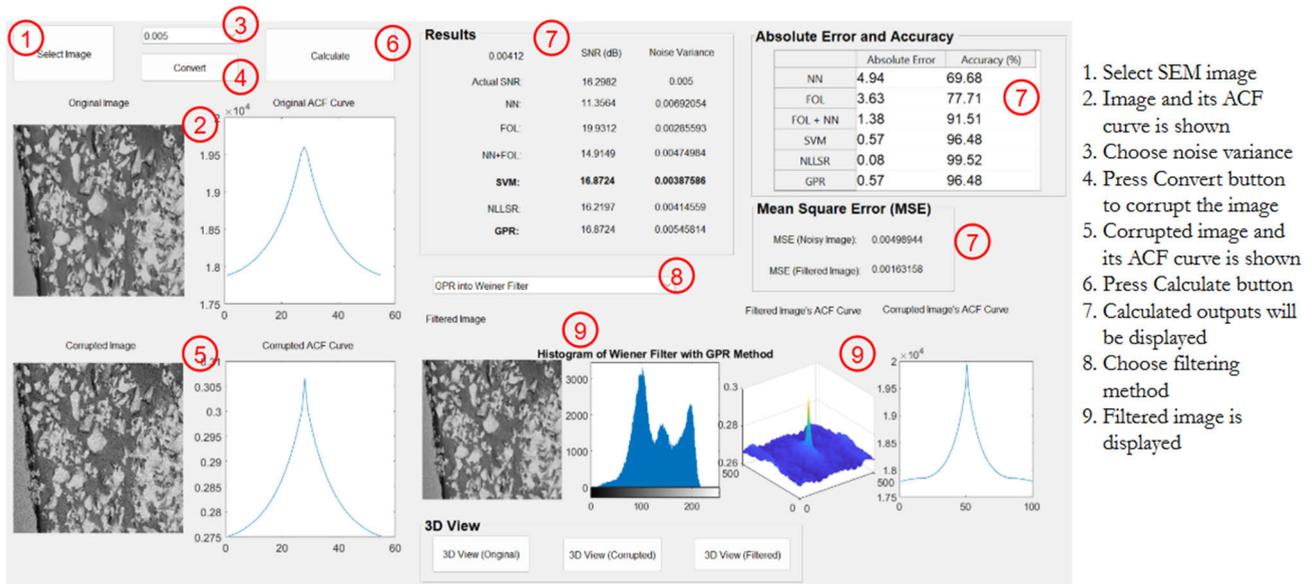

**FIGURE 11.** Graphical User Interface (GUI) developed for this research.

The interface connects noise measures with forecasting noise variations and Wiener filter operations into one tool that simplifies SEM image noise assessment and enhancement work. Through visual stage-by-stage inspection researchers can perform real-time evaluations regarding different filtering methods along with their noise reduction efficiency.

## IV. RESULTS AND DISCUSSION

The images used in developing SNR and Noise Variance (NV) estimation results were obtained from SE detector with the specifications as shown in Table 2.

**TABLE 2.** Machine settings for SEM images being used.

| Parameter | Value |
| --- | --- |
| Voltage gun | 10-30 KeV |
| Detector type | Secondary Emission detector |
| Horizontal field width | 10 $\mu m$ - 50 $\mu m$ |
| Magnification | 1000x-5000x |

### A. SNR ESTIMATION RESULT

Over 500 SEM images have been used to test the performance of SNR estimation methods, noise variance estimation methods and the proposed filter. Out of these 500 images, 400 images are used for training while the remaining 100 images are used for testing. Some examples of the images used are presented in Fig. 12. The SNR estimation methods being tested in this section include Nearest Neighborhood (NN), First-Order Linear (FOL) Interpolation, NN+FOL approach, Least Squares Regression (LSR), and Non-Linear Least Squares Regression (NLLSR). The actual SNR values serve as the ground truth, allowing for a direct comparison of the estimations from each method at different noise levels (0.001 to 0.01).

The comparison between the actual SNR and estimated SNR values for the 100 images in testing dataset (averaged) is presented in Table 3

**TABLE 3.** Comparison between actual SNR and estimated SNR (Averaged).

| Noise | Actual SNR | Estimated SNR | | | | |
| --- | --- | --- | --- | --- | --- | --- |
| | | NN | FOL | NN+FOL | LSR | NLLSR |
| 0.001 | 33.0461 | 17.5074 | 33.7588 | 25.8882 | 31.4261 | 28.2742 |
| 0.002 | 27.4855 | 15.9896 | 54.1365 | 22.5408 | 26.4658 | 24.1302 |
| 0.003 | 23.782 | 14.6837 | 32.0733 | 20.0955 | 23.3536 | 21.3465 |
| 0.004 | 21.6371 | 13.5988 | 26.7116 | 18.2947 | 21.1184 | 19.3397 |
| 0.005 | 19.6275 | 12.5911 | 23.4124 | 16.7715 | 19.2536 | 17.6101 |
| 0.006 | 17.9146 | 11.7113 | 20.8851 | 15.4479 | 17.7566 | 16.2114 |
| 0.007 | 16.447 | 10.9362 | 18.9877 | 14.3341 | 16.5904 | 15.0546 |
| 0.008 | 15.3248 | 10.176 | 17.5362 | 13.3544 | 15.5136 | 13.9767 |
| 0.009 | 14.4097 | 9.52379 | 16.229 | 12.4787 | 14.5224 | 13.0419 |
| 0.01 | 13.3683 | 8.83074 | 14.9726 | 11.5853 | 13.6445 | 12.1706 |

As can be seen in Table 3, SLR gives SNR values which are consistently closer to the actual SNR compared to other methods. This can be seen even clearly on Fig. 13, where the LSR's estimated SNR stays close to the actual SNR for all noise levels.

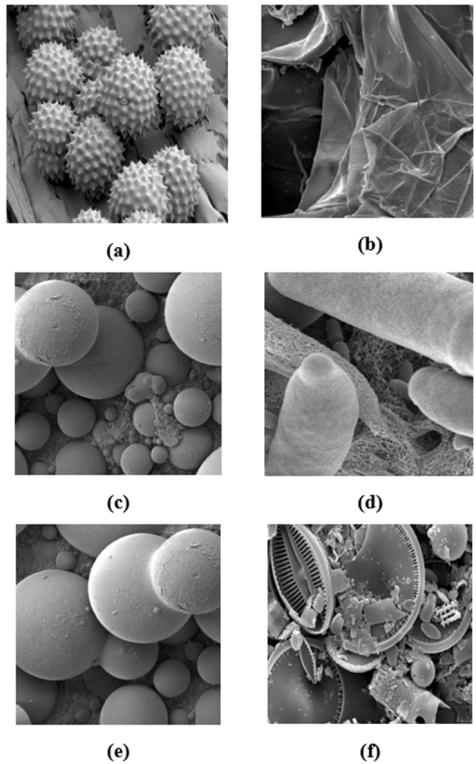

**FIGURE 12.** SEM sample images used in this study. (a) bacteria image, (b) wood fiber, (c) ball array composite image at low magnification, (d) bacteria image, (e) ball array composite image at high magnification, and (f) material electron microscope image.

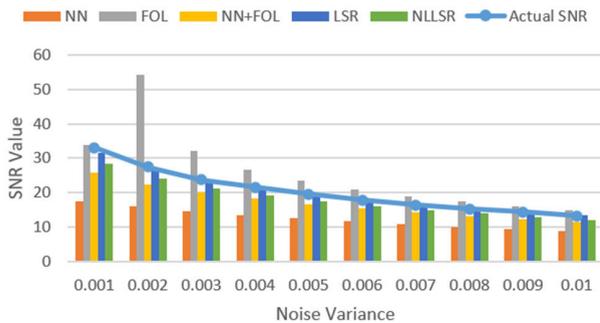

**FIGURE 13.** Comparison between estimated SNR and actual SNR.

### B. STATISTICAL TEST FOR SNR ESTIMATION RESULTS

In order to ensure the validity of the results presented in Table 3, a paired t-test is performed between the best performed method (LSR) against other methods with $\alpha = 0.05$. The test is based on the following hypotheses

$$H_0 : SNR_{LSR} \leq SNR_X \tag{40}$$
$$H_1 : SNR_{LSR} > SNR_X \tag{41}$$

where $X$ denotes the other method being compared. The results are shown in Table 4.

As seen from Table 4, every one-tailed comparison shows that $P(T \leq t) < 0.05$. Therefore, we fail to reject $H_0$, which means that we can safely assume that the results of the LSR outperforms other methods.

### C. NOISE VARIANCE ESTIMATION

This section presents a comparison of noise variance estimation using different SVM and GPR models. The same 500 SEM images were used as in SNR estimation, where 400 of them are used in training and the other 100 images are used for testing. The models being compared are Linear SVM, Quadratic SVM, Cubic SVM, Fine Gaussian SVM, Medium Gaussian SVM, Coarse Gaussian SVM, Optimizable SVM, Rational Quadratic GPR, Squared Exponential GPR, Matern 5/2 GPR, Exponential GPR and Optimizable GPR models.

Table 5 shows the SVM and GPR models comparison for training dataset, while Table 6 presents the same models being compared using testing dataset. From both tables, Root Mean Square Error (RMSE), Mean Square Error (MSE), $R^2$, and Mean Absolute Error (MAE) for each model is shown. A lower Root Mean Squared Error (RMSE) indicates that the model is more accurate and that its average forecast is closer to the actual values. Additionally, a high R-squared ($R^2$) value around 1 indicates a strong fit of the model since a larger percentage of the variance is explained by the model. A smaller Mean Squared Error (MSE) also indicates a smaller average squared difference between the anticipated and actual values. Lastly, a lower Mean Absolute Error (MAE) indicates the predicted values are very similar to the actual values on average, which means high accuracy [55].

The results shown that the Optimizable GPR is the best model in terms of RMSE, MSE, $R^2$ and MAE. This is because it has a lowest RMSE, a lowest MSE, $R^2$ value closest to 1 and a lowest MAE among the SVM and GPR models after trained. Among the SVM models, Optimizable SVM model has the best performance.

The detailed implementation for training the Optimizable SVM and GPR along with their parameters are shown in Table 7 and Table 8, respectively.

### D. FILTERING RESULTS

Finally, the noise variance estimated by the best model (AO-GPRLLSR) is inputted into a Wiener filter to form an adaptive Optimizable GPR LSR Wiener filtering method, which dynamically adjusts to the estimated noise characteristics to improve SEM image quality. Performance testing of the adaptive Wiener filter used Mean Squared Error (MSE) comparison between noisy images and their filtered versions to measure results. MSE functions as a measurement metric for image quality that decreases when the noise levels decrease and image quality increase. Images in this analysis came from the previous section about noise variance estimation and SNR estimation to maintain result consistency. The experimental design of the GPR evaluation based on MSE presented in this section is shown in Fig. 14.

**TABLE 4.** One-tailed t-test for SNR estimation ($\alpha$ =0.05).

| Metrics | LSR vs NN | | LSR vs FOL | | LSR vs NN+FOL | | LSR vs NLLSR | |
|---|---|---|---|---|---|---|---|---|
| | LSR | NN | LSR | FOL | LSR | NN+FOL | LSR | NLLSR |
| Mean | 0.48398 | 7.749397 | 0.48398 | 5.56606 | 0.48398 | 3.22515 | 0.48398 | 2.18867 |
| Variance | 0.231596 | 12.20013 | 0.231596 | 59.53868 | 0.231596 | 2.880845 | 0.231596 | 1.263146 |
| Observations | 10 | 10 | 10 | 10 | 10 | 10 | 10 | 10 |
| Pearson Correlation | 0.967327 | | 0.325871 | | 0.977304 | | 0.977181 | |
| Hypothesized Mean Difference | 0 | | 0 | | 0 | | 0 | |
| df | 9 | | 9 | | 9 | | 9 | |
| t Stat | -7.58308 | | -2.12214 | | -7.0405 | | -8.14822 | |
| P(T<=t) one-tail | 1.69E-05 | | 0.031409 | | 3.02E-05 | | 9.55E-06 | |
| t Critical one-tail | 1.833113 | | 1.833113 | | 1.833113 | | 1.833113 | |
| P(T<=t) two-tail | 3.39E-05 | | 0.062819 | | 6.05E-05 | | 1.91E-05 | |
| t Critical two-tail | 2.262157 | | 2.262157 | | 2.262157 | | 2.262157 | |

**TABLE 5.** SVM and GPR models comparison for training dataset.

| Model Type | RMSE | MSE | $R^2$ | MAE |
|---|---|---|---|---|
| Linear SVM | 0.007100597 | 5.04E-05 | 0.944655838 | 0.005209872 |
| Quadratic SVM | 0.007283188 | 5.30E-05 | 0.9417729 | 0.005232091 |
| Cubic SVM | 0.0065683 | 4.31E-05 | 0.95264258 | 0.0047831 |
| Fine Gaussian SVM | 0.006705683 | 4.50E-05 | 0.950640789 | 0.005061393 |
| Medium Gaussian SVM | 0.006505377 | 4.23E-05 | 0.953545579 | 0.004697819 |
| Coarse Gaussian SVM | 0.006600409 | 4.36E-05 | 0.952178433 | 0.004785519 |
| Optimizable SVM | 0.006476735 | 4.19E-05 | 0.953953737 | 0.004609295 |
| Rational Quadratic GPR | 0.006247356 | 3.90E-05 | 0.957157515 | 0.004340756 |
| Squared Exponential GPR | 0.006294805 | 3.96E-05 | 0.956504263 | 0.004379096 |
| Matern 5/2 GPR | 0.006213155 | 3.86E-05 | 0.957625321 | 0.004326175 |
| Exponential GPR | 0.006353306 | 4.04E-05 | 0.955692049 | 0.004505919 |
| Optimizable GPR | 0.006177919 | 3.82E-05 | 0.958104588 | 0.004305125 |

**TABLE 6.** SVM and GPR models comparison for testing dataset.

| Model Type | MAE | MSE | RMSE | $R^2$ |
|---|---|---|---|---|
| Linear SVM | 0.005143398 | 4.96E-05 | 0.007039994 | 0.945653253 |
| Quadratic SVM | 0.005138837 | 4.96E-05 | 0.007046189 | 0.945557572 |
| Cubic SVM | 0.004602176 | 4.06E-05 | 0.006370571 | 0.955497379 |
| Fine Gaussian SVM | 0.004467528 | 3.69E-05 | 0.006076623 | 0.959509462 |
| Medium Gaussian SVM | 0.004427803 | 4.05E-05 | 0.006363702 | 0.955593284 |
| Coarse Gaussian SVM | 0.004815231 | 4.28E-05 | 0.006541672 | 0.953074759 |
| Optimizable SVM | 0.0045982 | 4.14E-05 | 0.006431371 | 0.954643868 |
| Rational Quadratic GPR | 0.004176339 | 3.62E-05 | 0.006015802 | 0.960315951 |
| Squared Exponential GPR | 0.004197679 | 3.67E-05 | 0.006061668 | 0.959708517 |
| Matern 5/2 GPR | 0.004163059 | 3.58E-05 | 0.005986297 | 0.960704259 |
| Exponential GPR | 0.003905339 | 3.13E-05 | 0.005591654 | 0.965714582 |
| Optimizable GPR | 0.004128731 | 3.53E-05 | 0.005938075 | 0.961334799 |

The NV-guided Wiener filtering procedure leads to substantial MSE decreases which show in the Table 9. For each actual noise variance level, the filtered images consistently show a marked improvement in quality, as evidenced by the reduced MSE values.

The averaged results of the performance of the Wiener filter in terms of MSE before and after the filtering process is shown in Table 9.

Across all images in the dataset, the Optimizable GPR LSR Wiener filter shows effectiveness in reducing noise and improving image quality. The filtering process has consistently lowered MSE with better results at lower noise variance conditions while still maintains its filtering performance at higher noise variance conditions.

The improvement of MSE is also statistically significant, as shown from the one-tailed t-test we performed under the following hypotheses;

$$H_0 : MSE_{post\_filter} \leq MSE_{pre\_filter} \quad (42)$$

$$H_1 : MSE_{post\_filter} > MSE_{pre\_filter} \quad (43)$$

TABLE 7. Pseudocode for optimizable SVM training.

**Pseudocode 1** Optimizable SVM Training Phase
**Input:** Training data
**Output:** Trained SVM model, cross-validated RMSE
  1: **Extract predictors and response**
  2: **Set SVM parameters:**
    Kernel function: polynomial
    Polynomial order: 3
    Kernel scale: 1
    Box constraint: 0.0113
    Epsilon; 4.02e-5
    Standardize: true
  3: **Train the SVM model**
  4: **Create prediction function**
  5: **Perform 5-fold cross-validation**
  6: **Compute validation predictions and RMSE**
  **Return Trained GPR model, cross-validated RMSE**

TABLE 8. Pseudocode for optimizable GPR training.

**Pseudocode 2** Optimizable GPR Training Phase
**Input:** Training data
**Output:** Trained GPR model, cross-validated RMSE
  1: **Extract predictors and response**
  2: **Set GPR options (Bayesian Optimization output)**
    Basis function: none
    Kernel function: ardmatern32
    Sigma: 0.0338
    Standardize: false
  3: **Train the GPR model**
  4: **Create prediction function**
  5: **Perform 5-fold cross-validation**
  6: **Compute validation predictions and RMSE**
  **Return Trained GPR model, cross-validated RMSE**

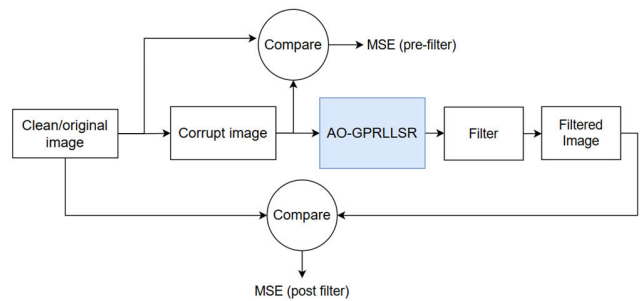

FIGURE 14. Filtering result evaluation pipeline.

The result is shown in Table 10.

From Table 7, we can conclude that since $P(T <= t) < \alpha$, the null hypothesis $H_0$ can be rejected.

To conclude the results, the analysis verifies Wiener filtering's effectiveness because the MSE reductions occur significantly at every tested noise variance. Next, the NV-guided Wiener filter achieves the best reduction ratios in environments with low noise variance condition because it operates most precisely at such noise levels. Under the high noise condition (0.01) the filter effectively denoises the images while achieving substantial MSE reductions relative to noisy images. The NV-guided Wiener filter exhibits adaptivity which enables it to tune based on the estimated noise values thus it can preserve important image features during noise filtering process. Lastly, the small variation of results in different images indicates that noise structure may slightly affect filtering efficiency. However, the overall image enhancement remains highly significant.

TABLE 9. Filter performance (Averaged).

| Actual NV | Filter performance | |
|---|---|---|
| | MSE (pre-filter) | MSE (post-filter) |
| 0.001 | 0.000987718 | 0.00045584 |
| 0.002 | 0.00196446 | 0.00077661 |
| 0.003 | 0.00293343 | 0.00106049 |
| 0.004 | 0.00386275 | 0.00130733 |
| 0.005 | 0.00484956 | 0.00156064 |
| 0.006 | 0.0057939 | 0.00179191 |
| 0.007 | 0.00671601 | 0.00200379 |
| 0.008 | 0.00764984 | 0.00220453 |
| 0.009 | 0.00859463 | 0.00241229 |
| 0.01 | 0.00949999 | 0.00257951 |

TABLE 10. One-tailed t-test for filter performance based on MSE ($\alpha$ =0.05).

| | MSE (post-filter) | MSE (pre-filter) |
|---|---|---|
| Mean | 0.001615294 | 0.005285 |
| Variance | 5.05289E-07 | 8.21E-06 |
| Observations | 10 | 10 |
| Pearson Correlation | 0.997195912 | |
| Hypothesized Mean Difference | 0 | |
| df | 9 | |
| t Stat | -5.381760183 | |
| P(T<=t) one-tail | 0.000221735 | |
| t Critical one-tail | 1.833112933 | |
| P(T<=t) two-tail | 0.000443471 | |
| t Critical two-tail | 2.262157163 | |

The findings demonstrate that the Optimizable GPR LSR Wiener filter represents an effective and adaptive solution for SEM image enhancement which proves useful in microscopy noise reduction tasks. After testing the filter performance, we use real-world SEM images to validate the filter performance.

### E. TESTS WITH ACTUAL IMAGES WITH UNKNOWN NOISE
The Optimizable GPR LSR Wiener filter is then validated based on the filter results of the real-world SEM images. We have used around 500 real-world SEM images with unknown noise levels to perform the validation. For each object in the image, four different copies of that image are generated under normal, low-speed, medium-speed, and high-speed scanning rate. The images obtained from normal scanning rate were used as the baseline, while higher scanning rate introduces higher noise. Effectively, for each object of scan, we obtain baseline/reference image, and three images with low, medium, and high noise. A series of tests were employed on the three noisy images (the non-baseline images) to test if the proposed method is able to predict the SNR and NV correctly which leads to filtered images which closely resemble the baseline images.

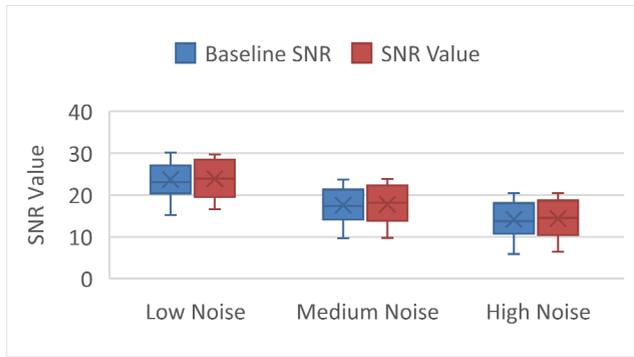

**FIGURE 15.** SNR of the baseline images vs estimated SNR of the same images in three categories of noise; low noise (low scanning speed), medium noise (medium scanning speed), and high noise (high scanning speed).

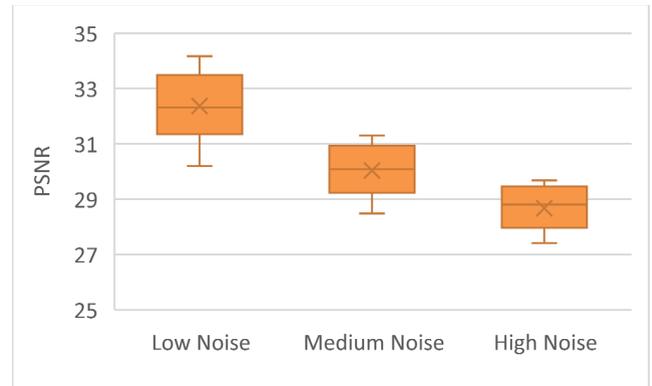

**FIGURE 17.** PSNR of filtered SEM images.

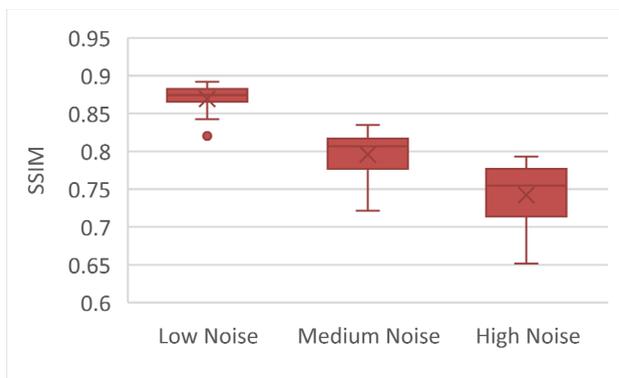

**FIGURE 16.** SSIM of filtered SEM images against the baseline images.

The first test is to see how close the estimated SNR is to the actual SNR of the noisy images, for all three noise level categories; low, medium, and high noise. In this test, we recorded the SNR values between the baseline images and the noisy images. As shown in Fig. 15, the estimated SNR stays close to the baseline SNR in all three noise levels. The LSR method demonstrates its ability to estimate SNR with minimal error at different noise intensity levels thus confirming its robustness.

In order to further test the similarity between baseline images and the noisy images which have been filtered with our proposed method and filtered using NV-guided Wiener Filter, Structural Similarity Index Measure (SSIM), which had been used for similar evaluation purpose in [56] is used. The result is presented in Fig. 16.

Structural Similarity Index (SSIM) is used to evaluate how well the filtering process maintains structural elements in images between the filtered SEM images and original SEM images. The SSIM results demonstrate that low-noise images produce the most excellent structural similarity measurements because they maintain values between 0.86 and 0.89 thus demonstrating a high level of accurate detail preservation during filtering. SSIM values measured from medium-noise images ranged between 0.72 and 0.83 while high-noise images documented the lowest SSIM values between 0.65 and 0.79. Higher amounts of noise coincide with a predictable decrease in structural similarity assessment results. The filter maintains excellent image structural preservation despite substantial noise during high-noise conditions by efficiently reducing noise in SEM images.

The next test is to check the Peak Signal to Noise Ratio (PSNR) to measure quality of the filtered images. The result is shown in Fig. 17.

The proposed filtering technique demonstrates high-performance in preserving original image quality different noise levels as shown by PSNR measurements. The PSNR values for low-noise images are between 30.2 to 34.17 and average at 32.2 which shows that the filter enhances images with minimal distortion. Next, medium-noise image evaluation based on the PSNR value shows a range from 28.48 to 31.31 while averaging to 29.9 as a balanced outcome of noise reduction and image clarity. The average PSNR values for high noise SEM images reached 28.7 while the measurements spanned from 27.41 to 29.68 due to the noise reduction capabilities. Even though the noise levels increase, PSNR reduces, but the filter delivers sustained acceptable outcomes because it excels at noise reduction while preserving important picture details.

The final test is the comparison against other filtering methods. In this test, the proposed AO-GPRLLSR with NV-guided Wiener Filter is compared against Median Filter, Gaussian Filter, and native Wiener Filter. The averaged MSE of the compared results are shown in Fig. 18.

As can be seen in Fig. 18, the NV-guided Wiener filter consistently produces lower MSE values which means that it outperforms other filtering techniques especially in higher noise levels.

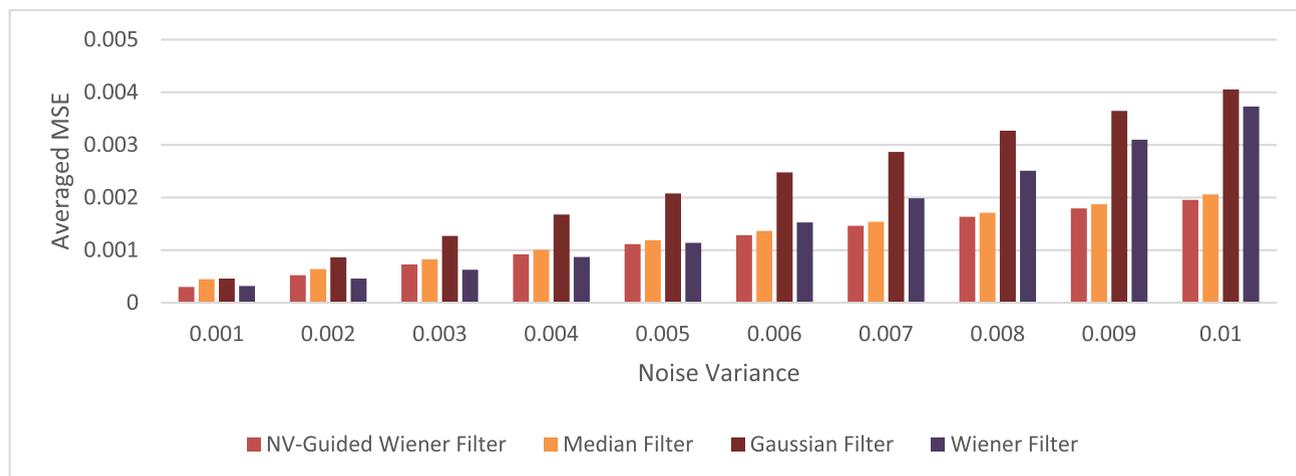

FIGURE 18. Comparison against other filtering techniques.

The AO-GPRLLSR in combination with NV-guided Wiener Filter demonstrates its successful ability to enhance SEM images through precise SNR estimation while simultaneously improving image. The accuracy of SNR estimation can be confirmed through its cosine similarity measurement which stays above 0.998 at every tested noise level. Besides, the PSNR metrics show that the filter successfully reduces noise from images while maintaining their quality although PSNR levels decrease when the noise levels rise. Lastly, the SSIM results show how filter operations maintain image structural details since low-noise images provide the best match to originals and medium-noise and high-noise images retain significant structural elements. The filter succeeds in retaining crucial image characteristics while noise increases because it produces small structural distortions. The proposed filtering method demonstrates reliability for SEM image enhancement based on the validated adaptability and robustness in recent results.

## V. CONCLUSION

The experimental research evaluated and proposed the Adaptive Optimizable Gaussian Process Regression Linear Least Squares Regression Filtering (AO-GPRLLSR) method for noise reduction in SEM imaging applications. The experimental results proved that this proposed technique eliminated white Gaussian noise while achieving better image quality. Integration of the Optimizable GPR model and LSR for SNR evaluation enabled the Wiener filter to adapt its filtering performance through Bayesian optimization noise variance estimation.

The validation results proved the successful operation of the method proposed. The calculated cosine similarities show outstanding precision in SNR measurement since they exceed 0.998 across every noise intensity. Besides, analysis of PSNR shows that the filter provides outstanding noise reduction together with retention of image quality where it exhibits higher PSNR figures during low-noise environments yet shows a diminishing pattern when noise increases. Next, the SSIM results demonstrate that the filter properly maintains important structural details under all conditions including those with high noise levels. Finally, comparison against median filter, gaussian filter, and native wiener filter also show that the proposed method produces cleaner images.

In conclusion, the experimental results demonstrate that AO-GPRLLSR operates as an efficient additive white Gaussian noise filter which enhances image visibility. The proposed method delivers powerful efficiency when removing additive white Gaussian noise which positions it as an attractive solution for all noise reduction applications. Future work should focus on developing real-time execution systems with imaging application-based data type adaptations and additional optimization techniques.

### ACKNOWLEDGMENT
The authors extend their appreciation to the creators and maintainers of the online dataset utilized in this study. Their invaluable contribution has significantly enriched their research.

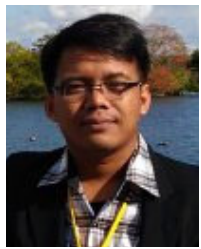

**IKSAN BUKHORI** received the bachelor's degree in control systems from President University, Indonesia, in 2013, and the Master of Philosophy degree in electronic system engineering from Universiti Teknologi Malaysia, Malaysia, in 2017. He is currently a Lecturer with the Electrical Engineering Department, Faculty of Engineering, President University. His research interests include control systems, robotics, artificial intelligence, and embedded systems.

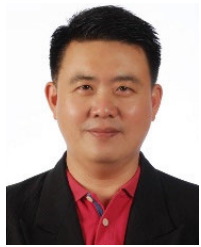

**KOK SWEE SIM** (Senior Member, IEEE) is currently a Professor with Multimedia University, Malaysia. He actively collaborates with various local and international universities and hospitals. He has filed 23 patents and 85 software copyrights. He is a fellow of the Academy of Sciences Malaysia, the Institution of Engineers Malaysia (IEM), and the Institution of Engineering and Technology (IET), U.K. Over the years, he has received numerous prestigious national and international awards. These include Japan Society for the Promotion of Science (JSPS) Fellowship, in 2018, the Top Research Scientists Malaysia (TRSM) Award from the Academy of Sciences Malaysia, in 2014, and Korean Innovation and Special Awards, in 2013, 2014, and 2015. He was a recipient of the TM Kristal Award and multiple World Summit on the Information Society (WSIS) Prizes, in 2017, 2018, 2019, 2020, and 2021.

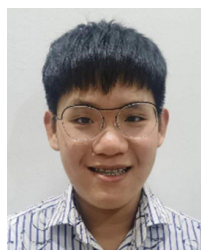

**DOMINIC CHEE YONG ONG** (Student Member, IEEE) received the Bachelor of Engineering degree (Hons.) in electronics majoring in robotics and automation from Multimedia University, Malaysia, in 2024, where he is currently pursuing the Master of Applied Engineering degree in IoT systems and technologies.

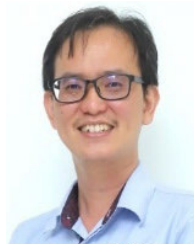

**KOK BENG GAN** received the Bachelor of Science degree (Hons.) in material physics from the Universiti Teknologi Malaysia, in 2001, and the Ph.D. degree in electrical, electronic, and system engineering from Universiti Kebangsaan Malaysia, in 2009. He was an Engineer in the field of electronic manufacturing services and original design manufacturing before venturing into academic research, in 2005. He is currently an Associate Professor with the Department of Electrical, Electronic and Systems Engineering, Faculty of Engineering and Built Environment, Universiti Kebangsaan Malaysia. He specializes in embedded systems and artificial intelligence in healthcare. His current research interests include biomedical optics and optical instrumentation, embedded systems and signal processing for medical applications, and biomechanics and human motion analysis.

• • •